\pdfoutput = 1
\documentclass[twocolumn]{article}

\usepackage{times}
\usepackage[numbers]{natbib}
\usepackage[english]{babel}
\usepackage{blindtext}
\usepackage{graphicx}
\usepackage{amsmath, amsthm, amssymb, bbm, bm}
\usepackage{enumerate}
\usepackage{float}      
\usepackage{subcaption}  
\usepackage{wrapfig}
\usepackage[margin=0cm]{caption}
\usepackage[titletoc, toc]{appendix}
\usepackage{tabularx}

\usepackage{multirow}
\usepackage{hhline}
\usepackage{makecell}
\usepackage{placeins}  

\usepackage[x11names, usenames, dvipsnames, svgnames, table]{xcolor}
\definecolor{firebrick}{rgb}{.698,.133,.133}
\definecolor{mybluelight}{rgb}{0.9, 0.9, 1.}
\definecolor{myorangelight}{rgb}{1., 0.9, 0.9}

\usepackage[utf8]{inputenc} 
\usepackage[T1]{fontenc}    
\usepackage{url}            
\usepackage{booktabs, colortbl}       
\usepackage{amsfonts}       
\usepackage{nicefrac}       
\usepackage{microtype}      

\usepackage{csquotes}
\usepackage{latexsym}

\usepackage{pifont}
\usepackage[boxruled, vlined, linesnumbered]{algorithm2e}
\SetAlFnt{\small}
\SetAlCapFnt{\small}
\SetAlCapNameFnt{\small}
\usepackage{algorithmic}
\algsetup{linenosize=\tiny}

\let\oldnl\nl
\newcommand{\nonl}{\renewcommand{\nl}{\let\nl\oldnl}}

\usepackage{paralist}

\usepackage{xspace}
\usepackage{soul}
\usepackage{dsfont}
\usepackage{stmaryrd}
\usepackage[textwidth=15mm]{todonotes}
\usepackage{dirtytalk}
\usepackage{pbox}
\usepackage{cprotect}

\usepackage{verbatim}
\usepackage{textcomp}
\usepackage[normalem]{ulem}

\usepackage{mathtools}
\usepackage{etextools}
\usepackage[inline]{enumitem}

\usepackage[colorlinks=true,allcolors=firebrick,bookmarks=false]{hyperref}

\definecolor{darkergreen}{RGB}{21, 152, 56}
\definecolor{red2}{RGB}{252, 54, 65}
\definecolor{Gray}{gray}{0.85}
\newcolumntype{g}{>{\columncolor{Gray}}c}

\newcommand\tableplus[1]{\textcolor{darkergreen}{#1}}

\let\OLDthebibliography\thebibliography
\renewcommand\thebibliography[1]{
  \OLDthebibliography{#1}
  \setlength{\parskip}{0pt}
  \setlength{\itemsep}{0pt plus 0.3ex}
}

\newcommand{\reals}{\mathbb{R}}

\newcommand{\abs}[1]{\ensuremath \left| #1 \right|}

\theoremstyle{definition}

\DeclarePairedDelimiterX{\divx}[2]{(}{)}{%
  #1\;\delimsize\|\;#2%
}

\newcommand*{\ie}{\emph{i.e.}\@\xspace}

\usepackage{style}

\newcommand\cl{\texttt{CL}\xspace}
\newcommand\corloc{\texttt{CorLoc}\xspace}

\newcommand\ytovone{\texttt{YTOv1}\xspace}
\newcommand\ytovtwodtwo{\texttt{YTOv2.2}\xspace}

\makeatletter
\@namedef{ver@everyshi.sty}{}
\newcommand{\removelatexerror}{\let\@latex@error\@gobble}

\title{TCAM: Temporal Class Activation Maps for Object Localization in \\
Weakly-Labeled Unconstrained Videos}

\renewcommand\footnotemark{}

\author{Soufiane~Belharbi$^{1}$,
  ~Ismail~Ben~Ayed$^{1}$,
  ~Luke~McCaffrey$^{2}$, and
  ~Eric~Granger$^{1}$\\
 	$^1$ LIVIA, Dept. of Systems Engineering, ÉTS, Montreal, Canada \\
	$^2$ Goodman Cancer Research Centre, Dept. of Oncology, McGill University, Montreal, Canada\\
{\tt\footnotesize \textcolor{black}{soufiane.belharbi.1@ens.etsmtl.ca} }
}

\newcommand{\ignore}[1]{}

\begin{document}
\maketitle\thispagestyle{fancy}

\maketitle
\rhead{\color{gray} \small Belharbi et al. \;  [WACV 2023]}

\begin{abstract}
Weakly supervised video object localization (WSVOL) allows locating object in videos using only global video tags such as object classes. State-of-art methods rely on multiple independent stages, where initial spatio-temporal proposals are generated using visual and motion cues, and then prominent objects are identified and refined. The localization involves solving an optimization problem over one or more videos, and video tags are typically used for video clustering. This process requires a model per video or per class making for costly inference. Moreover, localized regions are not necessary discriminant because these methods rely on unsupervised motion methods like optical flow, or discarded video tags from optimization. 
In this paper, we leverage the successful class activation mapping (CAM) methods, designed for WSOL based on still images. A new Temporal CAM (TCAM) method is introduced for training a discriminant deep learning (DL) model to exploit spatio-temporal information in videos, using an CAM-Temporal Max Pooling (CAM-TMP) aggregation mechanism over consecutive CAMs. In particular, activations of regions of interest (ROIs) are collected from CAMs produced by a pretrained CNN classifier, and generate pixel-wise pseudo-labels for training a decoder. In addition, a global unsupervised size constraint, and local constraint such as CRF are used to yield more accurate CAMs. Inference over single independent frames allows parallel processing of a clip of frames, and real-time localization.
Extensive experiments\footnote{Code: \href{https://github.com/sbelharbi/tcam-wsol-video}{https://github.com/sbelharbi/tcam-wsol-video}.} on two challenging YouTube-Objects datasets with unconstrained videos indicate that CAM methods (trained on independent frames) can yield decent localization accuracy. Our proposed TCAM method achieves a new state-of-art in WSVOL accuracy, and visual results suggest that it can be adapted for subsequent tasks, such as object detection and tracking.
\end{abstract}

\textbf{Keywords:} Convolutional Neural Networks, Weakly-Supervised Video Object Localization, Unconstrained Videos, Class Activation Maps (CAMs).
%
%

\begin{figure}[ht!]
\centering
  \centering
  \includegraphics[width=\linewidth]{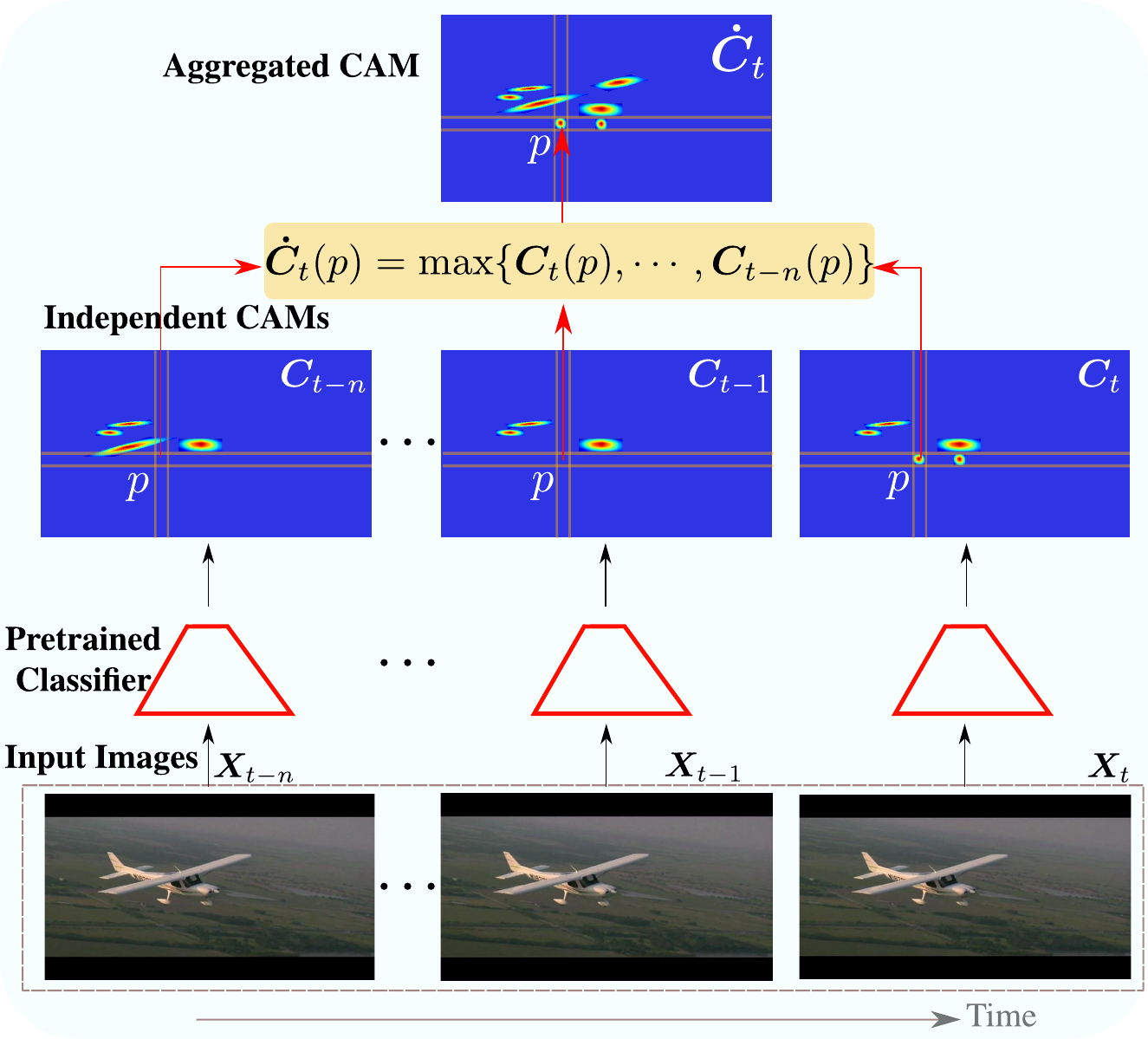}
  \caption{Example of CAM-Temporal Max Pooling (CAM-TMP) module for ROI aggregation of ${n+1}$ consecutive CAMs generated by a pretrained CNN classifier. It relies on the \emph{maximum} activation at location $p$ across the independent CAMs to produce the output CAM, ${\bm{\dot{C}}_t}$, that covers more discriminative parts. Notation is described in Sec.\ref{sec:method}.}
  \label{fig:cam-tmp}
\end{figure}

\begin{figure*}[ht!]
\centering
  \centering
  \includegraphics[width=0.7\linewidth]{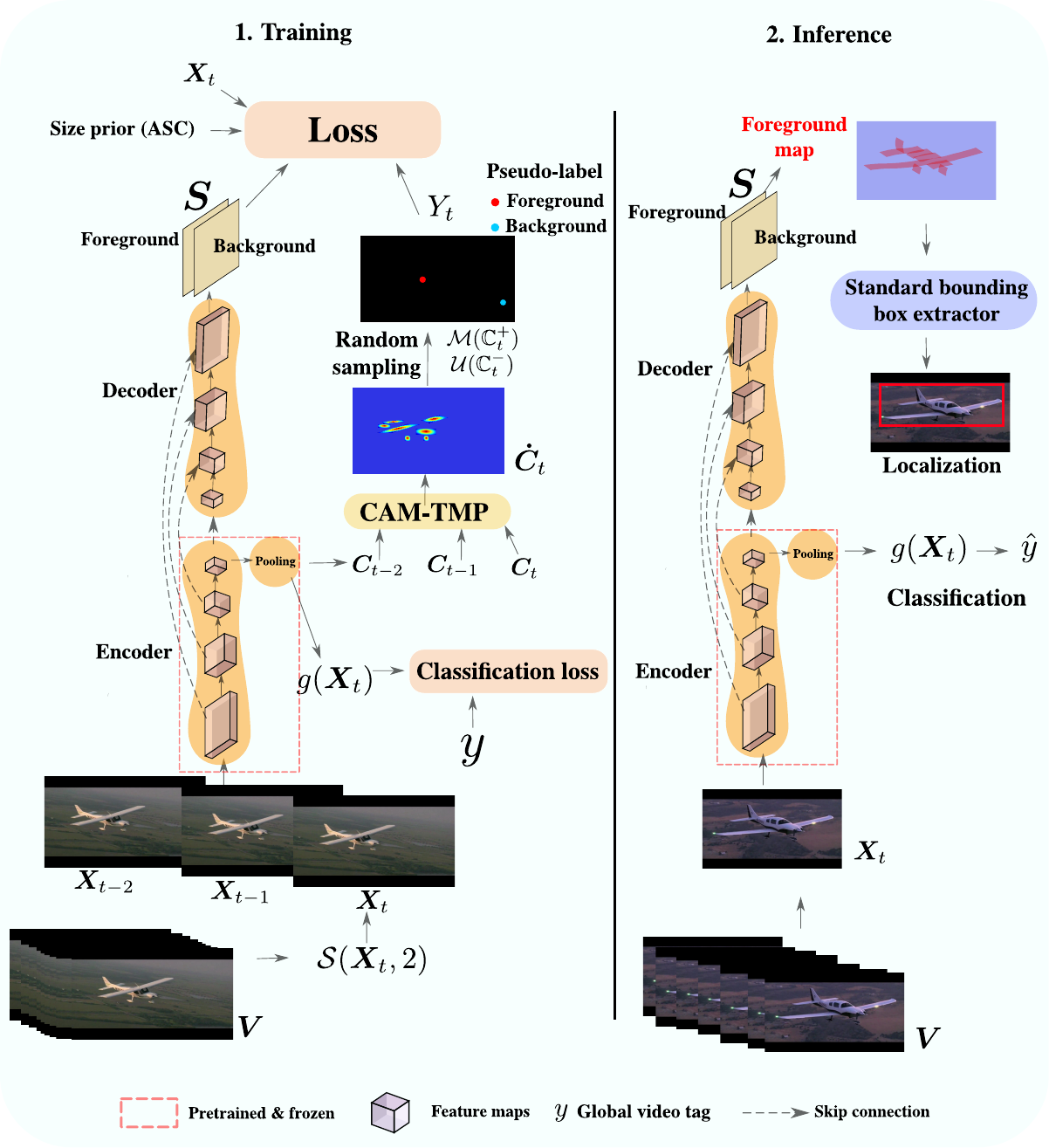}
  \caption{Our proposed TCAM method. \emph{Left}: training with temporal dependency $n=2$. \emph{Right}: inference (no temporal dependency). See notation in Sec.\ref{sec:method}.}
  \label{fig:proposal}
\end{figure*}

\section{Introduction}
\label{sec:intro}
A massive amount of videos can be easily accessed on the internet thanks to the rapid growth of video sharing platforms~\citep{ShaoCSJXYZX22,tang2013discriminative}. Therefore, the need to develop automatic methods to process and analyze these videos is of a great interest. The video object localization task plays a critical role toward video content understanding. It can improve the performance of subsequent tasks such as video summarization~\citep{ZhangHJYC17}, event detection~\citep{ChangYLZH16}, video object detection~\citep{Chen0HW20,HanWCQ20,ShaoCSJXYZX22}, and visual object tracking~\citep{BergmannML19,LuoXMZLK21}.

Videos are often captured in the wild with varying quality, and mostly without constraints (moving objects and camera, viewpoint changes, decoding artifacts, and editing effects). However, while being abundant, exploiting these videos for down-stream tasks is still an ongoing challenge, mainly due to the high cost of annotation. Compared to still images, labeling videos represents a more difficult and expensive process, as videos often contain a large number of frames. For the object localization task, bounding boxes are required for each frame. Given this cost, videos are typically weakly-labeled~\citep{JerripothulaCY16,tsai2016semantic} using class tags. A weak label is defined for the entire video, and often describes the main object or concept appearing in the video, without detailed spatio-temporal information. However, this translates to noisy/corrupted labels at frame level -- labels are assigned to an entire video, while only some of its frames may contain the object of interest.  Although using weak labels drastically reduce the cost of annotation, it creates additional challenges for visual recognition tasks like object localization. 

Weakly-supervised learning has emerged as an important paradigm to leverage coarse or global annotations like video tags, mitigating the need for bounding boxes annotation. Despite the importance of WSVOL, it has seen limited research~\citep{joulin2014,jun2016pod,Kwak2015,prest2012learning,RochanRBW16,zhang2020spftn}. Instead, the literature on weakly supervised video object segmentation (WSVOS) has dominated~\citep{Croitoru2019,FuXZL14,Halle2017,LiuTSRCB14,tsai2016semantic,Tokmakov2016,umer2021efficient,YanXCC17,ZhangJS14}, where bounding boxes are assumed to be produced via post-processing.

Most of the existing WSVOL methods are conventional, except for~\citep{Croitoru2019,Tokmakov2016,zhang2020spftn}. They usually generate spatio-temporal segments or proposals using visual and motion cues, and then prominent objects are identified and refined through post-processing. While they typically yield good performance, these methods present several limitations. They all require multiple sequential stages, and are not trained in an end-to-end manner. They are also costly at inference time since solutions are often optimized over a single video, or a cluster of videos from the same class. Moreover, they require building one model per class or per video, which is cumbersome in real-world applications, and scales poorly to a large number of classes. These methods are often non-discriminative -- video labels are not  \emph{explicitly} used in a differentiable way to extract objects, and instead only to cluster videos of same class. Consequentially, localized objects are not necessarily aligned \emph{semantically} with the video tag. Similarly, since almost all methods use motion cues, such as optical flow~\citep{LeeKG11,SundaramBK10}, they are prone to this alignment issue since such motion cues do not account for semantics. Lastly, motion information in unconstrained videos is very noisy due to movement of camera and objects.

%
To alleviate the aforementioned limitations, a new Temporal CAM (TCAM) method is proposed to train a single discriminative DL model through weakly-supervised learning. Our method requires only video tag annotations, and does not rely on additional assumptions. It is motivated by the success of Class Activation Mapping (CAM) methods~\citep{zhou2016learning}, applied for weakly-supervised object localization (WSOL) tasks on still images~\citep{belharbi2022fcam,ChoeS19,durand2017wildcat,gao2021tscam,JiangZHCW21layercam,SinghL17,wei2021shallowspol,zhou2016learning}. Using only global image-class labels, CAM methods allow training a DL model end-to-end, in a differentiable way, to localize discriminative image regions. Consequentially, localized ROIs are aligned with the semantic label of the image. At inference time, a CNN can rapidly classify an image and localized the corresponding ROI. This method scales well to a large number of classes, making it suitable for real-world applications.  However, these methods are limited to single images, and cannot exploit the temporal dependency between frames in videos. To leverage this spatio-temporal information, a new CAM Temporal Max-Pooling (CAM-TMP) mechanism is introduced to aggregate the ROIs from a sequence of CAMs (see Fig.\ref{fig:cam-tmp}). Our CAM-TMP simulates \emph{union} operation over ROIs in each frame by gathering ROIs from consecutive CAM, and thereby providing a better coverage of an object.

Our TCAM method relies on a U-Net style architecture~\citep{Ronneberger-unet-2015} to classify an image, and localize the corresponding ROI through full-resolution CAMs for better accuracy (see Fig.\ref{fig:proposal}). Using a pre-trained CNN to classify frames, our DL model is trained over a \emph{sequence} of successive frames, where CAM-TMP is used to accumulate ROIs. These are employed to generate reliable pseudo-labels for training the decoding architecture at the pixel-level. Following common practice~\citep{durand2017wildcat,zhou2016learning}, strong activations in a CAM are considered as foreground, while low activations are background. At each Stochastic Gradient Descent (SGD) step, we randomly sample foreground (FG) and background (BG) pixel pseudo-labels within independent CAMs~\citep{belharbi2022fcam,negevsbelharbi2022} to train the decoder. Such random sampling allows exploring FG/BG regions, and promotes the emergence of consistent CAMs. In contrast to standard CAMs for WSOL, our CAM-TMP generate activation maps that provide a better coverage of the true objects, which leads to better sampling of FG and BG pixels. To mitigate common issues of CAMs such as small ROI, we use unsupervised size prior~\citep{belharbi2022fcam,negevsbelharbi2022,belharbi2020minmaxuncer,pathak2015constrained} as a global constraints to encourage growth of both FG and BG regions, and avoid learning unbalanced CAMs. CRF loss~\citep{tang2018regularized} is also used to align CAMs with object boundaries by leveraging statistical properties of the image such as pixel color and proximity between pixels. Once the DL model is trained, inference is performed rapidly over independent frames, without considering temporal dependencies.  This is more suitable for real-time applications than other state-of-art WSVOL methods since TCAM is not required to process an entire video to localize within a single frame.

Our work aims to improve the state-of-art performance in WSVOL, while encouraging new research in this area.
\textbf{Our main contributions are summarized as follows.}

\noindent \textbf{(1)} We introduce the TCAM method, the first CAM-based method for WSVOL. In contrast with state-of-art WSVOL methods, TCAM allows training a single discriminative DL (U-Net style) model to process all classes at once. Our method is trained over unconstrained videos, each one annotated with a global class tag, and without any additional assumptions. Once trained, TCAM is able to rapidly predict the bounding box location for an object estimated based on any CAM method, along with the corresponding class label on each independent frame. 

\noindent \textbf{(2)} Unlike standard CAMs which are limited to still images, TCAM leverages spatio-temporal information in a sequence of CAMs. Using CAM-TMP module, we extract relevant ROIs from a sequence of generic CAMs provided by a pretrained CNN classifier. CAM-TMP then yields a single accurate CAM with better coverage of an object. Our loss exploits this CAM for training the decoder, by randomly sampling from its FG/BF regions. Additional constraint losses, including size prior and CRF, are used to obtained balanced and accurate CAMs. Note that our TCAM is generic, and can be integrated on top of any CAM method.

\noindent \textbf{(3)}  Extensive experiments conducted on two challenging public datasets -- YouTube-Objects v1.0 and v2.2 -- that are comprised of unconstrained videos indicate that: (a) standard CAM methods designed for WSOL on still images can achieve a high level of localization accuracy on frames from test set videos; (b) our TCAM method can achieves state-of-art in WSVOL accuracy. Results suggest that TCAM can be adapted for challenging downstream tasks, such as visual object detection and tracking.

\section{Related Work}
\label{sec:related-w}

\noindent \textbf{Weakly Supervised Video Object Localization.}
The limited amount of research in this category are often based on non-discriminative and non-deep models. These methods~\citep{jun2016pod,Kwak2015,prest2012learning,RochanRBW16,zhang2020spftn} initialize and select prominent proposals to be refined while considering spatio-temporal consistency constraints using mainly visual appearance and motion features. Different methods use prominent proposals as supervision to train a localizer~\citep{prest2012learning,zhang2020spftn}. Others rely on segmentation~\citep{RochanRBW16} followed by additional refinement using GrabCut~\citep{RotherKB04}. Single or cluster of videos are considered at once for optimization. 
For instance, POD method~\citep{jun2016pod} considers an iterative approach to localize a primary object in a video assumed to appear in most frames but not in all frames. It is achieved while empty frames are identified. Region proposals are initially generated via~\citep{AlexeDF12}. Each proposal is bisected into foreground and background. Models for foreground, background, and primary object are built. An iterative scheme is setup to refine each model in an evolutionary manner. The final primary model is used to select candidate proposals and locate the bounding box.
Recently, SPFTN~\citep{zhang2020spftn} considers a deep learning (DL) framework that jointly learns to segment and localize objects with noisy supervision estimation using advanced optical flow~\citep{LeeKG11}. Self-paced learning is considered to alleviate the ambiguity/noisy supervision.
Other methods~\citep{joulin2014} rely on co-localization of common object over a set of videos or images.

\noindent \textbf{Weakly Supervised Video Object Segmentation.} 
Most methods are non-deep and non-discriminative based models. They undergo multiple steps to perform segmentation. They often operate on a single video or a cluster of videos, and bounding boxes are obtained via post-processing. 

A set of methods initiate the learning by extracting independent spatio-temporal segments~\citep{HartmannGHTKMVERS12,tang2013discriminative,XuXC12,YanXCC17} using for instance unsupervised methods~\citep{BanicaAIS13,XuXC12} or generate proposals~\citep{zhang2015semantic} using pretrained detectors~\citep{zhang2015semantic}. These object-parts are then gathered using different features which mainly include visual appearance and motion cues, while preserving temporal consistency. This is often done using graph-based models such as Conditional random field (CRF) or GrabCut-like approach~\citep{RotherKB04}. DL models are rarely used.
For instance, authors in~\citep{LiuTSRCB14} propose a nearest neighbor-based label transfer between videos to deal with multi-class video segmentation. Videos are first segmented into spatio-temporal supervoxels~\citep{XuXC12} which then represented in high dimensional feature space using color, texture, and motion. A multi-video graph is built, and using appearance, this graph model encourages label smoothness between spatio-temporal adjacent supervoxels in the same video and supervoxels with similar appearance across other videos. This yields a final pixel segmentation.
M-CNN~\citep{Tokmakov2016} combines motion cues with a fully convolutional network (FCN). A Gaussian mixture model is used to estimate foreground appearance potentials via motion~\citep{papazoglou2013fast}. These potentials are combined with the FCN prediction to estimate label predictions of foreground using a GrabCut-like approach~\citep{RotherKB04}. These labels are used to fit the FCN. Authors use a fine-tuning stage over only few videos.

Other methods leverage co-segmentation to segment an object based on its occurrence on multiple images. A dominant approach is to use inter- and intra videos visual and motions cues to find common segments. Graphs, such as CRF and graph cuts, are used to model relations between variant segments~\citep{ChenCC12,FuXZL14,tsai2016semantic,ZhangJS14}.  For example, authors in~\citep{tsai2016semantic} generate object-like tracklets using a pretrained FCN. After collecting tracklets from all videos, they are linked for each object category via a graph. A sub-modular optimization is formulated to define the corresponding relation between tracklets based on their similarities while accounting for object properties such as appearance, shape, and motion. After maximizing this sub-modular function, tracklets are ranked using their mutual similarities allowing discovery of prominent objects in each video.

While all previous methods use labels to cluster videos of the same class, other methods do not use labels. However, the general process is roughly the same since previous methods do not exploit labels explicitly in their optimization. An initial guess of foreground regions is estimated~\citep{Croitoru2019,Halle2017,papazoglou2013fast,umer2021efficient}. This is achieved either using motion cues via~\citep{SundaramBK10} such as in~\citep{papazoglou2013fast}, Principal Component Analysis (PCA) such in~\citep{Halle2017,umer2021efficient}, or using VideoPCA algorithm~\citep{StretcuL15} as in~\citep{Croitoru2019}. This initial guess is not necessarily discriminative.
A final segmentation is then obtained by refinement using graphs~\citep{RotherKB04}. For instance, authors in~\citep{Croitoru2019} propose a DL model. It is based on an iterative learning process where at each iteration, a CNN teacher is trained to discover object in videos. Object discovery is achieved using VideoPCA algorithm~\citep{StretcuL15} which leverages spatio-temporal consistency in videos using appearance, shape, movement, and location of objects. Estimated foreground by the the teacher are fed to a CNN student for supervised training. Through iteration, several students are built and replace object discovery providing more reliable object segmentation.

\noindent \textbf{Weakly Supervised Object Localization in Still Images.}
Early work in WSOL~\citep{rony2022deep} focused on designing different spatial pooling layers, including Global Average Pooling (GAP)~\citep{lin2013network}, weighed GAP~\citep{zhou2016learning}, max-pooling~\citep{oquab2015object}, LSE~\citep{pinheiro2015image,sun2016pronet}, PRM~\citep{ZhouZYQJ18PRM}, WILDCAT ~\citep{durand2017wildcat,durand2016weldon}, and multi-instance learning pooling (MIL)~\citep{ilse2018attention}. However, these methods attained their limitation because CAMs can cover only small discriminative parts of the object. Subsequent work aimed to improve this aspect by refining the CAMs. This achieved through three different ways: 
\textit{(1) via data augmentation} by perturbating input image such as in HaS~\citep{SinghL17}, CutMix~\citep{YunHCOYC19}, AE~\citep{wei2017object}, ACoL~\citep{ZhangWF0H18}, MEIL~\citep{MaiYL20eil}, and MaxMin~\citep{belharbi2020minmaxuncer}; or by perturbating features as in SPN~\citep{zhu2017soft}, GAIN~\citep{LiWPE018CVPR}, and ADL~\citep{ChoeS19}, 
or \textit{(2) via architectural changes} such as in NL-CCAM~\citep{YangKKK20}, FickleNet~\citep{LeeKLLY19}, DANet~\citep{XueLWJJY19iccvdanet}, I${^2}$C~\citep{ZhangW020i2c}, ICL~\citep{KiU0B20icl}, and TS-CAM~\citep{gao2021tscam},
or \textit{(3) by using pseudo-labels for fine-tuning} such as in SPG~\citep{ZhangWKYH18}, PSOL~\citep{ZhangCW20rethink}, SPOL~\citep{wei2021shallowspol}, FCAM~\citep{belharbi2022fcam}, NEGEV~\citep{negevsbelharbi2022}, and DiPS~\citep{murtaza2022dips,murtaza2022dipssypo}. Other methods aim to produce the bounding box directly without CAMs~\citep{MeethalPBG20icprcstn}. 
All the aforementioned methods extract localization from forward pass in the model. Other methods rely on forward and backward pass to estimate CAMs. This includes methods that  
\textit{(1) are biologically inspired} such as feedback layer~\citep{cao2015look}, and Excitation-backprop~\citep{zhang2018top}, 
or \textit{(2) rely on gradient-aggregation} such as GradCAM~\citep{SelvarajuCDVPB17iccvgradcam}, GradCam++~\citep{ChattopadhyaySH18wacvgradcampp},  XGradCAM~\citep{fu2020axiom}, and LayerCAM~\citep{JiangZHCW21layercam}, 
or \textit{(3) use confident-aggregation } to avoid gradient saturation~\citep{Adebayo2018,Kindermans2019} such as Ablation-CAM~\citep{desai2020ablation}, Score-CAM~\citep{WangWDYZDMH20scorecam}, SS-CAM~\citep{naidu2020sscam}, and IS-CAM~\citep{naidu2020iscam}. Despite the success of these methods, they are limited to still images and they are not equipped to leverage temporal information in videos.

Our proposal benefits from the simplicity of CAM methods, which provide single discriminative DL model for the WSOL task, to mitigate several issues in WSVOL. In addition, our TCAM method leverages the spatio-temporal information in videos.

\section{Proposed Approach}
\label{sec:method}

\noindent \textbf{Notation.} Let ${\mathbb{D} = \{(\bm{V}, y)_i\}_{i=1}^N}$ denotes a training set of videos, where ${\bm{V} = \{\bm{X}_t\}_{t=1}^{t=T}}$ is an input video with $T$ frames, and ${\bm{X}_t: \Omega \subset \reals^2}$ is the $t$th frame in the video; ${y \in \{1, \cdots, K\}}$ is the video global class label, with $K$ the number of classes, and ${\Omega}$ is a discrete image domain. Assume that all frames inherit the same class label as the video global class tag.  Our model is a U-Net style architecture~\citep{Ronneberger-unet-2015} (Fig.\ref{fig:proposal}). It is composed of two parts: (a) classification module ${g}$ with parameters ${\bm{W}}$. It performs image classification. (b) segmentation module (decoder) ${f}$ with parameters ${\bm{\theta}}$. It outputs two CAMs, one for the foreground and the other for background. The classifier ${g}$ is composed of an encoder backbone for building features, and a pooling head to yield classification scores. We denote by ${g(\bm{X}) \in [0, 1]^K}$ the per-class classification probabilities where ${g(\bm{X})_k = \mbox{Pr}(k | \bm{X})}$. The classifier ${g}$ is trained using standard cross-entropy to correctly classify independent frames,
    ${\min_{\bm{W}} \;  - \log(\bm{g}(\bm{X})[y]).}$

Once trained, its weights ${\bm{W}}$ are frozen, and not considered for future training. Classifier ${g}$ can yield a CAM of the target ${y}$, referred to as ${\bm{C}}$. We note ${\bm{C}_t}$ as the corresponding CAM of the the frame ${\bm{X}_t}$ at time ${t}$. The decoder generates softmax activation maps denoted as ${\bm{S}_t = f(\bm{X}_t) \in [0, 1]^{\abs{\Omega} \times 2}}$. Note that ${\bm{S}^0_t, \bm{S}^1_t}$ refer to the background and foreground maps, respectively.  Let ${\bm{S}_t(p) \in [0, 1]^2}$ denotes a row of matrix ${\bm{S}_t}$, with index ${p \in \Omega}$ indicating a point within ${\Omega}$. The operation ${\mathcal{S}(\cdot, n)}$ provides the set of ${n}$ previous neighbors of an element in the same video, plus the element itself. For instance, ${\mathcal{S}(\bm{X}_t, n) = \{\bm{X}_t, \bm{X}_{t-1}, \cdots, \bm{X}_{t-n}\}}$ is the set of $n$ previous frames of the frame ${\bm{X}_t}$,  and ${\mathcal{S}(\bm{C}_t, n) = \{\bm{C}_t, \bm{C}_{t-1}, \cdots, \bm{C}_{t-n}\}}$ is the set of $n$ previous CAMs of the CAM ${\bm{C}_t}$.

\noindent \textbf{CAM Temporal Max-Pooling (CAM-TMP).} A sequence of frames in a video often captures the same scene with minimal variations. Therefore, objects within the scene have small displacement. However, this small change can cause CAMs to vary slightly, and highlight different minimal parts of the object as ROI. We leverage this behavior to build a single CAM, ${\bm{\dot{C}}_{t}}$, that covers more parts at once. This CAM will be used later for sampling foreground/background regions. To this end, we propose an aggregation method between consecutive CAMs where we perform a \emph{union} operation between all spotted ROI in each CAM. This is achieved by taking the \emph{maximum} CAM activation through time over activation in the same position of a sequence of CAMs (Fig.\ref{fig:cam-tmp}). This is similar to spatial max-pooling operation, commonly used in CNNs, that seeks the presence of the object in small spatial neighborhood. Our temporal max-pooling seeks to determine whether one of the CAMs has activated over the object in a sequence of CAMs. At spatial position $p$, we formulate our CAM-TMP by taking the maximum across all CAMs at the same position,
\begin{equation}
    \label{eq:eq1}
    \bm{\dot{C}}_t(p) = \max \{ \bm{C}^1(p), \cdots, \bm{C}^{n+1}(p) \}, \; \bm{C}^i \in \mathcal{S}(\bm{C}_t, n) ,
\end{equation}
where ${\bm{C}^i}$ is the $i$th element of the set ${\mathcal{S}(\bm{C}_t, n)}$, and $p \in \Omega$.

\noindent \textbf{Sampling Pseudo-Labels.} To guide the training of the decoder $f$, we exploit pixel-wise pseudo-supervision collected from the previously built CAM ${\bm{\dot{C}}_t}$ for the corresponding frame ${\bm{X}_t}$. We rely on the common assumption that strong activations in a CAM are more likely to be foreground, and low activations are considered background~\citep{belharbi2022fcam,negevsbelharbi2022,durand2017wildcat,zhou2016learning}.
We denote ${\mathbb{C}^+_t}$ as foreground region, estimated via the operation ${\mathcal{O}^+}$. It is determined as pixels with activations greater than Otsu threshold~\citep{otsuthresh} estimated over ${\bm{\dot{C}}_t}$. The leftover region, estimated via the operation ${\mathcal{O}^-}$, is considered more likely background ${\mathbb{C}^-_t}$. 
\begin{equation}
    \label{eq:sets}
    \mathbb{C}^+_t = \mathcal{O}^+(\bm{\dot{C}}_t), \quad \mathbb{C}^-_t = \mathcal{O}^-(\bm{\dot{C}}_t) \;.
\end{equation}
Both foreground and background regions are noisy and uncertain. The region ${\mathbb{C}^-_t}$ is still likely to contain part of the object. Similarly,
${\mathbb{C}^+_t}$ may still contain background. Due to this uncertainty, we avoid to directly fit these regions to the model. Instead, we consider a stochastic sampling over each region to avoid overfitting and allow the emergence of consistent regions~\citep{belharbi2022fcam,negevsbelharbi2022}. For each frame, and at each SGD step, we randomly select one pixel as foreground pseudo-label, and another single pixel as background pseudo-label. Their location is represented in,
\begin{equation}
    \label{eq:sset}
    \Omega^{\prime}_t = \mathcal{M}(\mathbb{C}^+_t)\; \cup \; \mathcal{U}(\mathbb{C}^-_t) \;, 
\end{equation}
where ${\mathcal{M}(\mathbb{C}^+_t)}$ is a multinomial sampling distribution function over foreground region that samples a single location using the magnitude of pixels activations located exclusively in ${\mathbb{C}^+_t}$. Therefore, strong activation are more likely to be sampled as foreground.

Uniform sampling distribution ${\mathcal{U}(\mathbb{C}^-_t)}$ is used to sample a single background pixel from ${\mathbb{C}^-_t}$. We favor uniform random exploration of background region since the background is evenly distributed over the image. However, foreground region is only distributed over one place where the object is located. We denote by ${Y_t}$ the \emph{partially} pseudo-labeled mask for the sample ${\bm{X}_t}$, where ${Y_t(p) \in \{0, 1\}^2}$ with labels ${0}$ for background, and ${1}$ for foreground. This mask holds the sampled locations in Eq. \ref{eq:sset} and their pseudo-labels. Locations with unknown pseudo-labels are encoded as unknown.

\noindent \textbf{Overall Training Loss.} Our training loss considers a frame ${\bm{X}_t}$ and its $n$ previous frames, \ie ${\mathcal{S}(\bm{X}_t, n)}$, to leverage spatio-temporal information in a video. The time position $t$ is uniformly and randomly sampled in the video. The loss is composed of three parts.
\textbf{a)} pixel-wise alignment using pseudo-label ${Y_t}$. This is achieved using partial cross-entropy,
\begin{equation}
    \label{eq:eqh}
    \begin{aligned}
    &\bm{H}_p(Y_t, \bm{S}_t) =\\ 
    &- (1 - Y_t(p))\; \log(1 - \bm{S}_t^0(p))- Y_t(p) \; \log(\bm{S}_t^1(p)) \;.
    \end{aligned}
\end{equation}
\textbf{b)} To avoid the common unbalanced problem CAMs ${\bm{S}_t}$, where the background dominates the foreground (or the opposite), a global constraint is considered -- the absolute size constraint (ASC)~\citep{belharbi2020minmaxuncer} over both regions. We do not assume whether the background is larger than the foreground~\citep{pathak2015constrained} nor the opposite. This constraint pushes both regions to be large, and it is formulated as inequality constraints which are then solved via a standard log-barrier method~\citep{boyd2004convex}. 
\textbf{c)} To avoid trivial solution of ASC, where half the image is foreground and the other half is background, we use an additional local term that leverage pixels statistics including color and proximity. In particular, the CRF loss~\citep{tang2018regularized}, denoted by ${\mathcal{R}}$ is included to ensure that CAM activations are consistent with the object boundaries and sampling regions. 

Our overall total loss is formulated as,
\begin{equation}
\label{eq:totalloss}
\begin{aligned}
\min_{\bm{\theta}} \quad & \sum_{p \in \Omega^{\prime}_t} \bm{H}_p(Y_t, \bm{S}_t) + \lambda\; \mathcal{R}(\bm{S}_t, \bm{X}_t) \;,\\
\textrm{s.t.} \quad & \sum \bm{S}^r_t \geq 0 \;, \quad r \in \{0, 1\} \;,\\
\end{aligned}
\end{equation}
where ${\sum \bm{S}^0_t, \sum \bm{S}^1_t}$ are the area size of the background and foreground regions, respectively.

The training of our method (Eq. \ref{eq:totalloss}) only requires the video global tag $y$ to train the classifier ${g}$, and properly estimate the pseudo-label mask ${Y_t}$ corresponding to the correct object class labeled in the video. This ensures that the semantic meaning of the foreground in ${\bm{S}^1_t}$ is aligned with the true label ${y}$. The spatio-temporal dependency between CAMs is leveraged in Eq. \ref{eq:eq1} to compute ${\bm{\dot{C}}_t}$ which is then used to sample ${Y_t}$.
Our final trained model is evaluated on single independent frames, thereby producing a class prediction for the object in the fame, along with its spatial localization ${\bm{S}^1_t}$. Therefore, frames can be processed in parallel saving more inference time. Standard methods may be used for bounding box estimation in CAMs (Fig.\ref{fig:proposal}, \emph{right})~\citep{choe2020evaluating}.

\begin{table*}
\begin{center}
\resizebox{\linewidth}{!}{
\begin{tabular}{|c|l|*{10}{c|}|g|c|}
\hline
\textbf{Dataset} & \textbf{Method \emph{(venue)}} & \textbf{Aero} & \textbf{Bird} & \textbf{Boat} & \textbf{Car} & \textbf{Cat} & \textbf{Cow} & \textbf{Dog} & \textbf{Horse} & \textbf{Mbike} & \textbf{Train} & \textbf{Avg} & \textbf{Time/Frame} \\
\hhline{-----------||---}
\noalign{\vspace{\doublerulesep}}
\hhline{-----------||---}
 &\citep{prest2012learning} {\small \emph{(cvpr,2012)}} & 51.7 & 17.5 & 34.4 & 34.7 & 22.3 & 17.9 & 13.5 & 26.7 & 41.2 & 25.0 & 28.5 & N/A   \\
 &\citep{papazoglou2013fast} {\small \emph{(iccv,2013)}} & 65.4 & 67.3 & 38.9 & 65.2 & 46.3 & 40.2 & 65.3 & 48.4 & 39.0 & 25.0 & 50.1 & 4s  \\
 &\citep{joulin2014} {\small \emph{(eccv,2014)}} & 25.1 & 31.2 & 27.8 & 38.5 & 41.2 & 28.4 & 33.9 & 35.6 & 23.1 & 25.0 & 31.0 & N/A \\
 &\citep{Kwak2015} {\small \emph{(iccv,2015)}}  & 56.5 & 66.4 & 58.0 & 76.8 & 39.9 & 69.3 & 50.4 & 56.3 & 53.0 & 31.0 & 55.7 & N/A  \\
 &\citep{RochanRBW16} {\small \emph{(ivc,2016)}} & 60.8 & 54.6 & 34.7 & 57.4 & 19.2 & 42.1 & 35.8 & 30.4 & 11.7 & 11.4 & 35.8 & N/A \\
 &\citep{Tokmakov2016} {\small \emph{(eccv,2016)}} & 71.5 & 74.0 & 44.8 & 72.3 & 52.0 & 46.4 & 71.9 & 54.6 & 45.9 & 32.1 & 56.6 & N/A  \\
 &POD~\citep{jun2016pod} {\small \emph{(cvpr,2016)}} & 64.3 & 63.2 & 73.3 & 68.9 & 44.4 & 62.5 & 71.4 & 52.3 & 78.6 & 23.1 & 60.2 & N/A  \\
 &\citep{tsai2016semantic} {\small \emph{(eccv,2016)}} & 66.1 & 59.8 & 63.1 & 72.5 & 54.0 & 64.9 & 66.2 & 50.6 & 39.3 & 42.5 & 57.9 & N/A  \\
 &\citep{Halle2017} {\small \emph{(iccv,2017)}} & 76.3 & 71.4 & 65.0 & 58.9 & 68.0 & 55.9 & 70.6 & 33.3 & 69.7 & 42.4 & 61.1 & 0.35s  \\
&\citep{Croitoru2019} (LowRes-Net\textsubscript{iter1}) {\small \emph{(ijcv,2019)}} & 77.0 & 67.5 & 77.2 & 68.4 & 54.5 & 68.3 & 72.0 & 56.7 & 44.1 & 34.9 & 62.1 & 0.02s  \\
\multirow{6}{*}{\ytovone } & \citep{Croitoru2019} (LowRes-Net\textsubscript{iter2}) {\small \emph{(ijcv,2019)}} & 79.7 & 67.5 & 68.3 & 69.6 & 59.4 & 75.0 & 78.7 & 48.3 & 48.5 & 39.5 & 63.5 & 0.02s\\
&\citep{Croitoru2019} (DilateU-Net\textsubscript{iter2}) {\small \emph{(ijcv,2019)}} & 85.1 & 72.7 & 76.2 & 68.4 & 59.4 & 76.7 & 77.3 & 46.7 & 48.5 & 46.5 & 65.8 & 0.02s \\
&\citep{Croitoru2019} (MultiSelect-Net\textsubscript{iter2}) {\small \emph{(ijcv,2019)}} & 84.7 & 72.7 & 78.2 & 69.6 & 60.4 & 80.0 & 78.7 & 51.7 & 50.0 & 46.5 & 67.3 & 0.15s \\
&SPFTN (M)~\citep{zhang2020spftn} {\small \emph{(tpami,2020)}} & 66.4 & 73.8 & 63.3 & 83.4 & 54.5 & 58.9 & 61.3 & 45.4 & 55.5 & 30.1 & 59.3 & N/A  \\
&SPFTN (P)~\citep{zhang2020spftn} {\small \emph{(tpami,2020)}}& \textbf{97.3} & 27.8 & \textbf{81.1} & 65.1 & 56.6 & 72.5 & 59.5 & \textbf{81.8} & 79.4 & 22.1 & 64.3 & N/A  \\
&FPPVOS~\citep{umer2021efficient} {\small \emph{(optik,2021)}} & 77.0 & 72.3 & 64.7 & 67.4 & 79.2 & 58.3 & 74.7 & 45.2 & \textbf{80.4} & 42.6 & 65.8 & 0.29s  \\
\cline{2-14}
&CAM~\citep{zhou2016learning} {\small \emph{(cvpr,2016)}} & 75.0 & 55.5 & 43.2 & 69.7 & 33.3 & 52.4 & 32.4 & 74.2 & 14.8 & 50.0 & 50.1 & 0.2ms  \\
&GradCAM~\citep{SelvarajuCDVPB17iccvgradcam} {\small \emph{(iccv,2017)}} & 86.9 & 63.0 & 51.3 & 81.8 & 45.4 & 62.0 & 37.8 & 67.7 & 18.5 & 50.0 & 56.4 & 27.8ms  \\
&GradCAM++~\citep{ChattopadhyaySH18wacvgradcampp} {\small \emph{(wacv,2018)}} & 79.8 & 85.1 & 37.8 & 81.8 & 75.7 & 52.4 & 64.9 & 64.5 & 33.3 & \textbf{56.2} & 63.2 & 28.0ms  \\
&Smooth-GradCAM++~\citep{omeiza2019corr} {\small \emph{(corr,2019)}} & 78.6 & 59.2 & 56.7 & 60.6 & 42.4 & 61.9 & 56.7 & 64.5 & 40.7 & 50.0 & 57.1 & 136.2ms  \\
&XGradCAM~\citep{fu2020axiom} {\small \emph{(bmvc,2020)}} & 79.8 & 70.4 & 54.0 & \textbf{87.8} & 33.3 & 52.4 & 37.8 & 64.5 & 37.0 & 50.0 & 56.7 & 14.2ms  \\
&LayerCAM~\citep{JiangZHCW21layercam} {\small \emph{(ieee,2021)}} & 85.7 & \textbf{88.9} & 45.9 & 78.8 & 75.5 & 61.9 & 64.9 & 64.5 & 33.3 & \textbf{56.2} & 65.6 & 17.9ms  \\
&TCAM (ours) & 90.5 & 70.4 & 62.2 & 75.7 & \textbf{84.8} & \textbf{81.0} & \textbf{81.0} & 64.5 & 70.4 & 50.0 & \textbf{73.0} & 18.5ms  \\
\hhline{-----------||---}
\noalign{\vspace{1.5mm}}
\hhline{-----------||---}
&\citep{Halle2017}  {\small \emph{(iccv,2017)}}& 76.3 & 68.5 & 54.5 & 50.4 & 59.8 & 42.4 & 53.5 & 30.0 & 53.5 & \textbf{60.7} & 54.9 & 0.35s   \\
&\citep{Croitoru2019} (LowRes-Net\textsubscript{iter1}) {\small \emph{(ijcv,2019)}} & 75.7 & 56.0 & 52.7 & 57.3 & 46.9 & 57.0 & 48.9 & 44.0 & 27.2 & 56.2 & 52.2 & 0.02s    \\
&\citep{Croitoru2019} (LowRes-Net\textsubscript{iter2}) {\small \emph{(ijcv,2019)}} & 78.1 & 51.8 & 49.0 & 60.5 & 44.8 & 62.3 & 52.9 & 48.9 & 30.6 & 54.6 &  53.4 & 0.02s    \\
&\citep{Croitoru2019} (DilateU-Net\textsubscript{iter2}){\small \emph{(ijcv,2019)}} & 74.9 & 50.7 & 50.7 & 60.9 & 45.7 & 60.1 & 54.4 & 42.9 & 30.6 & 57.8 & 52.9 & 0.02s   \\
\multirow{6}{*}{\ytovtwodtwo } & \citep{Croitoru2019} (BasicU-Net\textsubscript{iter2}){\small \emph{(ijcv,2019)}} & \textbf{82.2} & 51.8 & 51.5 & 62.0 & 50.9 & 64.8 & 55.5 & 45.7 & 35.3 & 55.9 & 55.6 & 0.02s  \\
&\citep{Croitoru2019} (MultiSelect-Net\textsubscript{iter2}){\small \emph{(ijcv,2019)}} & 81.7 & 51.5 & 54.1 & 62.5 & 49.7 & 68.8 & 55.9 & 50.4 & 33.3 & 57.0 & 56.5 & 0.15s  \\
\cline{2-14}
&CAM~\citep{zhou2016learning} {\small \emph{(cvpr,2016)}} & 52.3 & 66.4 & 25.0 & 66.4 & 39.7 & \textbf{87.8} & 34.7 & 53.6 & 45.4 & 43.7 & 51.5 & 0.2ms  \\
&GradCAM~\citep{SelvarajuCDVPB17iccvgradcam} {\small \emph{(iccv,2017)}} & 44.1 & 68.4 & 50.0 & 61.1 & 51.8 & 79.3 & 56.0 & 47.0 & 44.8 & 42.4 & 54.5 & 27.8ms  \\
&GradCAM++~\citep{ChattopadhyaySH18wacvgradcampp} {\small \emph{(wacv,2018)}} & 74.7 & 78.1 & 38.2 & 69.7 & 56.7 & 84.3 & 61.6 & 61.9 & 43.0 & 44.3 & 61.2 & 28.0ms  \\
&Smooth-GradCAM++~\citep{omeiza2019corr} {\small \emph{(corr,2019)}} & 74.1 & 83.2 & 38.2 & 64.2 & 49.6 & 82.1 & 57.3 & 52.0 & 51.1 & 42.4 & 59.5 & 136.2ms  \\
&XGradCAM~\citep{fu2020axiom} {\small \emph{(bmvc,2020)}} & 68.2 & 44.5 & 45.8 & 64.0 & 46.8 & 86.4 & 44.0 & 57.0 & 44.9 & 45.0 & 54.6 & 14.2ms  \\
&LayerCAM~\citep{JiangZHCW21layercam} {\small \emph{(ieee,2021)}} & 80.0 & 84.5 & 47.2 & \textbf{73.5} & 55.3 & 83.6 & 71.3 & 60.8 & 55.7 & 48.1 & 66.0 & 17.9ms  \\
&TCAM (ours) & 79.4 & \textbf{94.9} & \textbf{75.7} & 61.7 & \textbf{68.8} & 87.1 & \textbf{75.0} & \textbf{62.4} & \textbf{72.1} & 45.0 & \textbf{72.2} & 18.5ms  \\
\hline
\end{tabular}}
\end{center}
\caption{Localization performance (\corloc) on the \ytovone~\citep{prest2012learning} and \ytovtwodtwo~\citep{KalogeitonFS16} test sets.}
\label{tab:yto}
\end{table*}

\section{Results and Discussion}
\label{sec:results}

\subsection{Experimental Methodology}
\label{subsec:exp-method}

\noindent \textbf{Datasets.} For evaluation, we perform experiments on unconstrained video datasets for WSVOL task where videos are labeled globally using class label for training, and frame bounding boxes are provided for evaluation. In particular, we consider two public challenging datasets: YouTube-Object v1.0 (\ytovone~\citep{prest2012learning}) and v2.2 (\ytovtwodtwo~\citep{KalogeitonFS16} ) datasets. We follow common protocol for WSVOL task~\citep{KalogeitonFS16,prest2012learning}.

\textit{YouTube-Object v1.0 (\ytovone) \citep{prest2012learning}:} This dataset is composed of videos collected from YouTube\footnote{\url{https://www.youtube.com}} by querying for the names of 10 object classes. Each class has between 9 and 24 videos with a duration that varies from 30 seconds to 3 minutes. It contains 155 videos where each video is split into short duration clips named shots. There are 5507 shots, with each gathering multiple frames, reaching in total 571 089 frames. In each shot, only few frames are annotated with a bounding box to localized object of interest. The authors divided the dataset into 27 testing videos with a total of 396 labeled bounding boxes, and 128 video for training. It is common to use part of the training videos as validation set. In our experiments, we consider 5 random videos per class which amounts to a total of 50 videos for validation.

\textit{YouTube-Object v2.2 (\ytovtwodtwo)~\citep{KalogeitonFS16}:} This is an extension and improvement of \ytovone. It contains more frames, 722 040 frames in total. More importantly, authors provided more bounding boxes annotations. They divided the dataset into 106 videos for training, and 49 videos for test. For validation set, we consider, in our case, 3 random videos per class from the trainset. Compared to \ytovone, the test set contains much more annotation. It holds 1 781 frames with bounding boxes annotation, and  a total of 2 667  bounding boxes. This makes this release much more challenging.

\noindent \textbf{Evaluation Measure.} For localization performance, \corloc metric~\citep{deselaers2012weakly} is used. It represents the percentage of predicted bounding boxes that have an Intersection Over Union (IoU) between prediction and ground truth greater than half (IoU > 50\%). In addition, standard classification accuracy, \cl, is used to measure classification performance. It is measured over frames with bounding boxes.

\noindent \textbf{Implementations Details.} 
In all our experiments, we train for 100 epochs with 32 mini-batch size. Following WSOL task~\citep{choe2020evaluating}, we used ResNet50~\citep{heZRS16} as a backbone. Images are resized to ${256\times256}$, then randomly cropped to ${224\times224}$ for training. The temporal dependency ${n}$ in Eq.\ref{eq:eq1} is set via the validation set from the set ${n \in \{1, \cdots, 10\}}$.
In Eq.\ref{eq:totalloss}, the hyper-parameter $\lambda$ for the CRF is set to the same value as in \citep{tang2018regularized} that is ${2e^{-9}}$.  For log-barrier optimization, hyper-parameter $t$ is set to the same value as in \citep{belharbi2019unimoconstraints,kervadec2019log}. It is initialized to $1$, and increased by a factor of ${1.01}$ in each epoch with a maximum value of $10$. In all experiments, we used a learning rate in ${\{0.1, 0.01, 0.001\}}$ using SGD for optimization. Our classifier is pretrained on independent frames. In all our experiments, and due to the large number of redundant frames per video, we randomly select a different frame in each shot at each epoch. This allows training CNNs over videos in a reasonable time.

\noindent \textbf{Baseline Methods.} For validation, we compare our method to available results. In particular, we compare to~\citep{
Croitoru2019, Halle2017, joulin2014, Kwak2015, papazoglou2013fast, prest2012learning, RochanRBW16, Tokmakov2016, tsai2016semantic}, POD~\citep{jun2016pod}, SPFTN~\citep{zhang2020spftn}, and FPPVOS~\citep{umer2021efficient}. Additionally, we implemented several CAM-based methods for further comparison. This includes CAM~\citep{zhou2016learning}, GradCAM~\citep{SelvarajuCDVPB17iccvgradcam}, GradCam++~\citep{ChattopadhyaySH18wacvgradcampp}, Smooth-GradCAM++~\citep{omeiza2019corr}, XGradCAM~\citep{fu2020axiom}, and LayerCAM~\citep{JiangZHCW21layercam}. CAM-based methods are trained on independent frames. In all our experiments, we use LayerCAM~\citep{JiangZHCW21layercam} to generate CAMs used to build a complete CAM  ${\bm{\dot{C}}_t(p)}$ (Eq.\ref{eq:eq1}), which is then used to build pseudo-labels ${Y_t}$ (Eq.\ref{eq:totalloss}). Note that our method is generic. It can be used with any CAM  method.

\subsection{Results}
\label{subsec:compare}

\noindent \textbf{Comparison with State-of-Art\footnote{The supplementary material provides some additional results, demonstrative videos.}.}
Tab. \ref{tab:yto} presents the results obtained on both datasets \ytovone, and \ytovtwodtwo. We first note that CAM-based methods are very competitive compare to previous state-of-the-art methods. In particular, GradCAM++~\citep{ChattopadhyaySH18wacvgradcampp} and LayerCAM~\citep{JiangZHCW21layercam}  achieved an average localization performance of \corloc of ${63.1\%, 65.6\%}$ over \ytovone, and ${61.2\%, 66.0\%}$ over \ytovtwodtwo, respectively. Previous state-of-the-art methods yielded ${67.3\%}$, and ${56.5\%}$, respectively. This demonstrates the benefit of discriminative training of CAMs even though they are not aware of temporal dependency.
Training our CAM-based method with temporal awareness between frames has boosted the localization performance furthermore reaching new state-of-the-art results.
The same table shows as well that all methods suffer a discrepancy in performance between different objects where some classes are easier than others. For instance, the class 'train' seems very difficult.  In CAM-based methods, we noticed that 'train' localization is often mistaken with railway track since they often co-occur together. In addition, this object is often filmed from close range, at stations, leading to a large object, that often covers the entire frame making its localization difficult.

Although it is not commonly provided, we present classification performance in Tab.\ref{tab:cl-perf} for our trained CAM-methods since they are able to do both tasks: classification, and localization. These methods yielded descent classification performance. However, there is a large margin between both datasets showing the difficulty of \ytovtwodtwo dataset.

{
\setlength{\tabcolsep}{3pt}
\renewcommand{\arraystretch}{1.1}
\begin{table}[ht!]
\centering
\resizebox{\linewidth}{!}{%
\centering
\small
\begin{tabular}{lcgcg}
\textbf{Methods} & &  \ytovone && \ytovtwodtwo  \\
\cline{1-1}\cline{3-3} \cline{5-5} \\
CAM~\citep{zhou2016learning} {\small \emph{(cvpr,2016)}} &  &                     85.3 && 73.9  \\
GradCAM~\citep{SelvarajuCDVPB17iccvgradcam} {\small \emph{(iccv,2017)}} &  &                     85.3 && 71.3  \\
GradCAM++~\citep{ChattopadhyaySH18wacvgradcampp} {\small \emph{(wacv,2018)}} &  &                     84.4 && 72.4  \\
Smooth-GradCAM++~\citep{omeiza2019corr} {\small \emph{(corr,2019)}} &  &                     82.6 && \textbf{75.2}  \\
XGradCAM~\citep{fu2020axiom} {\small \emph{(bmvc,2020)}} &  &                     \textbf{87.3} && 71.6  \\
LayerCAM~\citep{JiangZHCW21layercam} {\small \emph{(ieee,2021)}} &  &                     84.4 && 72.1  \\
TCAM (ours) &  &                     84.4 && 72.1  \\
\cline{1-1}\cline{3-3} \cline{5-5} \\
\end{tabular}
}
\caption{Classification accuracy (\cl) on test set of \ytovone~\citep{prest2012learning} and \ytovtwodtwo~\citep{KalogeitonFS16} datasets.}
\label{tab:cl-perf}
\vspace{-1em}
\end{table}
}

\noindent \textbf{Ablation Studies.}
We performed an ablation study for key components of our loss function, using LayerCAM~\citep{JiangZHCW21layercam} as baseline to generate CAMs for pseudo-labels (see Tab.\ref{tab:ablation-parts}). We observe that using only pseudo-labels improved the localization accuracy from ${65.6\%}$ to ${68.5\%}$. Adding CRF helps localization, but using only size constraint did not provide much  benefits compared to the baseline alone. Combining pseudo-labels, CRF, and size constraint yielded the best localization performance of ${70.5\%}$ but without considering temporal dependency. Adding our temporal module, CAM-TMP, improves the localization accuracy up to ${73\%}$, indicating its benefit.

\begin{figure}[ht!]
\centering
  \centering
  \includegraphics[width=\linewidth]{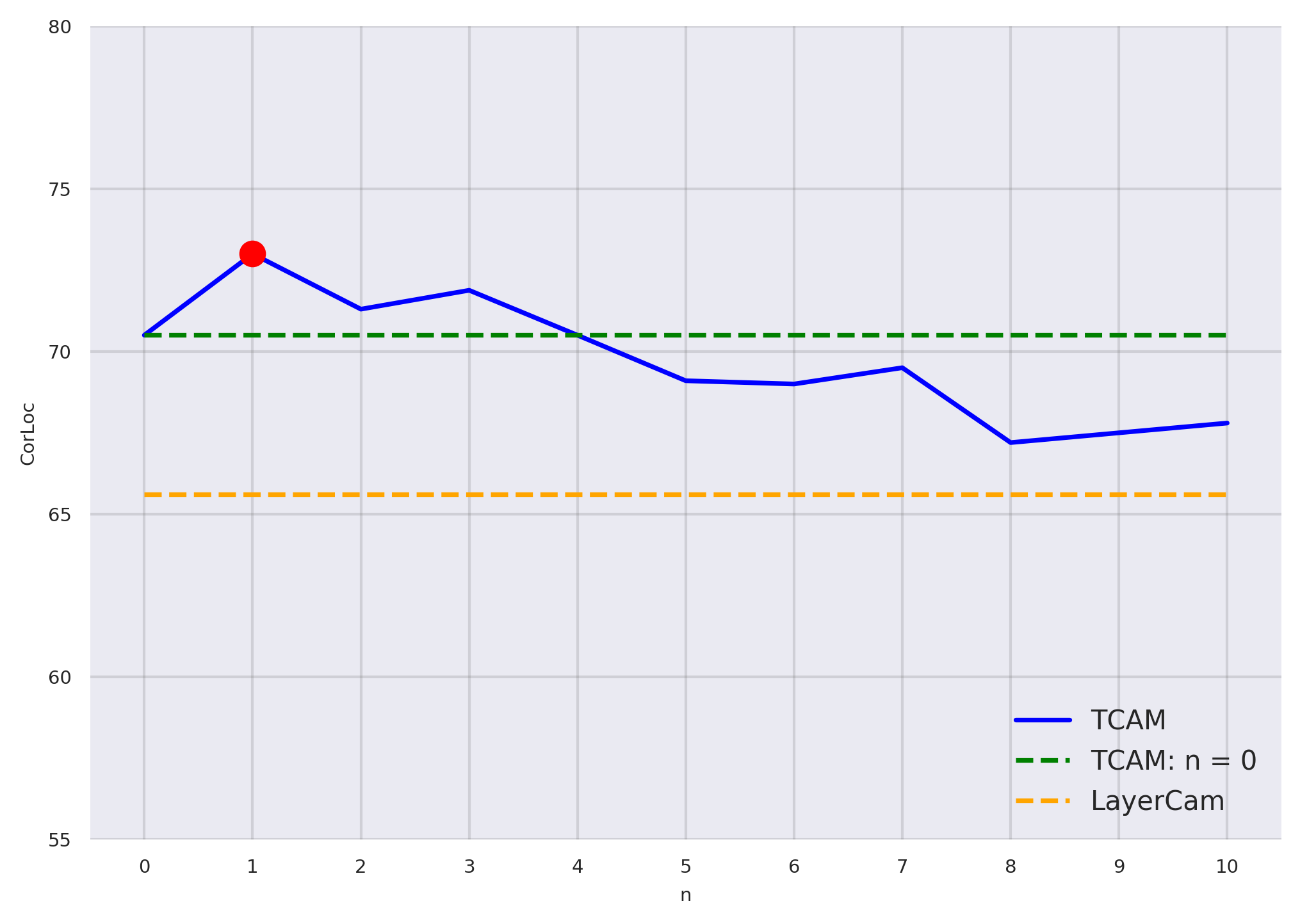}
  \caption{Localization accuracy (\corloc) of TCAM with different temporal dependencies $n$ on the \ytovone test set.}
  \label{fig:ablation-range-time}
\end{figure}

In addition, the impact of time range dependency is investigated (see Fig.\ref{fig:ablation-range-time}). As expected, considering previous frames ($n > 0$) helped in improving localization compared to looking only to instant frames (${n=0}$). However, long range dependency hampers the performance after ${n=1}$. After ${n=4}$, localization performance drops below the case of ${n=0}$. This is thought to be caused by \emph{object displacement}. Spatial locations in nearby frames cover typically the same objects. Therefore, ROI in CAMs are expected to land on same objects. Consequently, collecting ROI via Eq.\ref{eq:eq1} is expected to be beneficial. However, moving to far away frames makes the same spatial location cover \emph{different} objects, therefore collecting the wrong object. As a result, while our proposed module, \ie CAM-TMP, can leverage temporal dependency in videos to improve localization, it is limited to short range frames. Nonetheless, using long range time dependency still yields better performance than the baseline method LayerCAM~\citep{JiangZHCW21layercam} (Fig.\ref{fig:ablation-range-time}). Based on our results, we recommend using short range dependency. We mention that such factor is strongly tied to the video frame rate. Using long range dependency in fast frame rate could be safe. However, slow frame rate should be considered with caution. We note that the information of video frame rate is not provided in the studied datasets.

{
\setlength{\tabcolsep}{3pt}
\renewcommand{\arraystretch}{1.1}
\begin{table}[ht!]
\centering
\resizebox{\linewidth}{!}{%
\centering
\small
\begin{tabular}{lllcg}
\multicolumn{3}{l}{\textbf{Methods} }& &  \corloc  \\
\cline{1-3}\cline{5-5} \\
\multicolumn{3}{l}{Layer-CAM~\citep{JiangZHCW21layercam} {\small \emph{(ieee,2021)}}} &  &                     \tableplus{65.6}  \\
\cline{1-3}\cline{5-5} \\
\multirow{3}{*}{$n=0$} & &       
Ours + $\mathbb{C}^+_t$ + $\mathbb{C}^-_t$ &  &   $68.5$       \\
&& Ours + $\mathbb{C}^+_t$ + $\mathbb{C}^-_t$ + CRF &  &   $69.6$       \\
&& Ours + $\mathbb{C}^+_t$ + $\mathbb{C}^-_t$ + ASC &  &   $66.2$      \\
&& Ours + $\mathbb{C}^+_t$ + $\mathbb{C}^-_t$ + CRF + ASC &  &   $70.5$\\
\cline{1-1}\cline{3-3} \cline{5-5} \\
$n>0$ & & Ours + $\mathbb{C}^+_t$ + $\mathbb{C}^-_t$ + CRF + ASC + CAM-TMP &  &   $73.0$ \\
\cline{1-1}\cline{3-3} \cline{5-5} \\
\multicolumn{3}{l}{Improvement} &  &                                      \tableplus{+7.4}\\
\cline{1-3}\cline{5-5} \\
\end{tabular}
}
\caption{Localization accuracy (\corloc) of TCAM with different losses on the \ytovone test set.}
\label{tab:ablation-parts}
\vspace{-1em}
\end{table}
}

\begin{figure}
     \centering
     \begin{subfigure}[b]{0.45\textwidth}
         \centering
         \includegraphics[width=\textwidth]{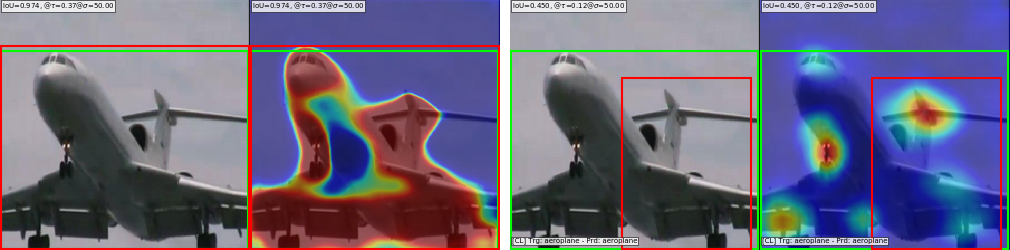}
     \end{subfigure}
     \\
     \begin{subfigure}[b]{0.45\textwidth}
         \centering
         \includegraphics[width=\textwidth]{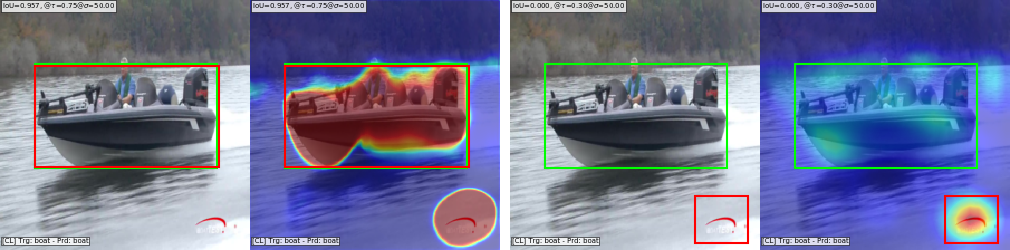}
     \end{subfigure}
     \begin{subfigure}[b]{0.45\textwidth}
         \centering
         \includegraphics[width=\textwidth]{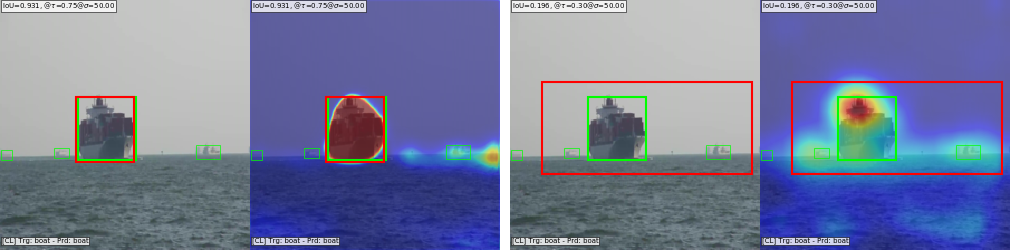}
     \end{subfigure}
     \\
     \begin{subfigure}[b]{0.45\textwidth}
         \centering
         \includegraphics[width=\textwidth]{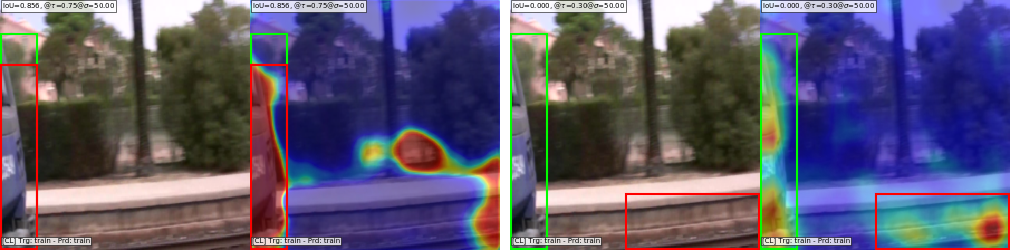}
     \end{subfigure}
      \\
     \begin{subfigure}[b]{0.45\textwidth}
         \centering
         \includegraphics[width=\textwidth]{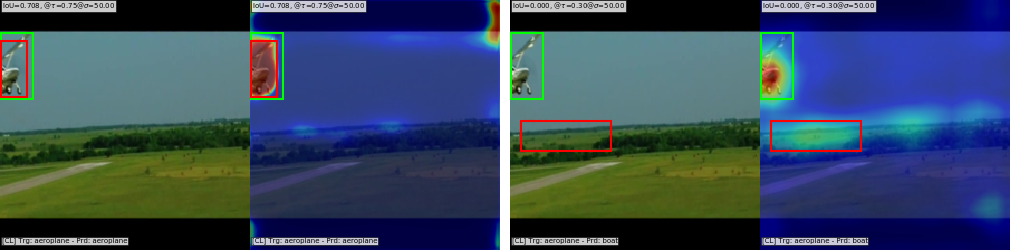}
     \end{subfigure}
     \\
     \begin{subfigure}[b]{0.45\textwidth}
         \centering
         \includegraphics[width=\textwidth]{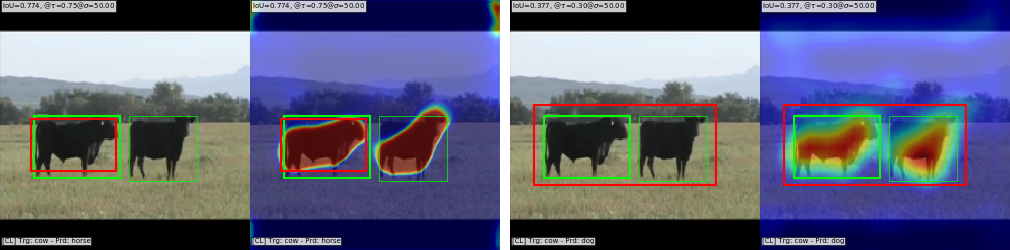}
     \end{subfigure}
        \caption{Prediction examples of test sets frames. \emph{Left}: TCAM (ours). \emph{Right}: baseline CAM method, LayerCAM~\citep{JiangZHCW21layercam}. \emph{Bounding box}: ground truth (green), prediction (red). Second column is predicted CAM over image.}
        \label{fig:pred}
\end{figure}

\begin{figure}
     \centering
     \begin{subfigure}[b]{0.45\textwidth}
         \centering
         \includegraphics[width=\textwidth]{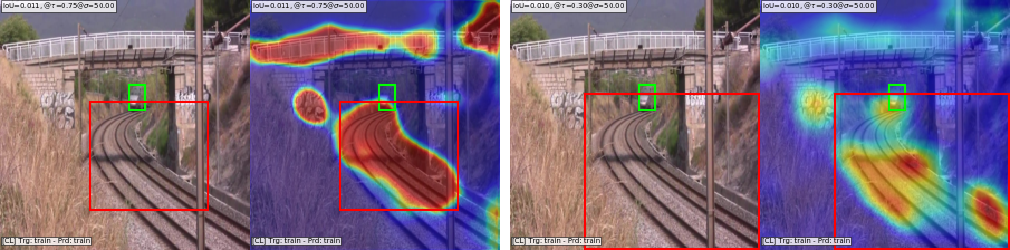}
     \end{subfigure}
     \\
     \begin{subfigure}[b]{0.45\textwidth}
         \centering
         \includegraphics[width=\textwidth]{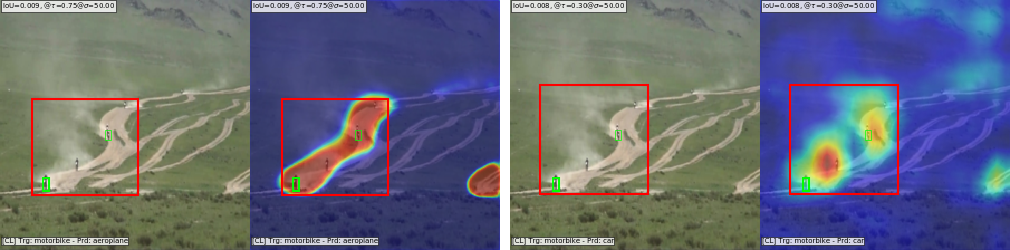}
     \end{subfigure}
        \caption{Typical failed cases of our method over test sets. \emph{Left}: TCAM (ours). \emph{Right}: baseline CAM method,  LayerCAM~\citep{JiangZHCW21layercam}. \emph{Bounding box}: ground truth (green), prediction (red). Second column is predicted CAM over image.}
        \label{fig:failure}
\end{figure}

\noindent \textbf{Visual Results.}
Fig.\ref{fig:pred} illustrates prediction cases over labeled ground truth frames. Our method yields CAMs that tend to cover the entire object with a clear distinction between foreground and background. It deals well with multi-instances, and partially visible objects. The second row shows a concrete case where random sampling prevents overfitting over small and strong ROI (bottom-right), and allows other consistent and discriminative objects to emerge (boat at center) from low activations. Fig.\ref{fig:failure} shows typical failure of our method. They manifest by over-activation over tiny objects, and spill to background. This is mainly caused by heavy presence of wrong ROI activations over the baseline CAM used to generated pseudo-labels. Unfortunately, dominant erroneous pseudo-labels in this case can lead to wrong localization in our method. Their early detection in the baseline CAM and dealing with them is of paramount importance for future extensions of this work. Such issues go under learning with noisy labels which is still an ongoing active domain~\citep{songsurvey2020}. This highlights the dependency of our method to the accuracy of the backbone CNN classifier, and baseline CAM. We provide in supplementary material demonstrative videos, in addition to our code.

\noindent \textbf{Demonstrative Videos.}
For demonstration, we provide illustrative videos for localization using our method. Videos can be found on this Google drive link: \href{https://drive.google.com/drive/folders/1D8DgOdjT35Vf5Tqej3K5ZWFqz3LhgeQt?usp=sharing}{https://drive.google.com/drive/folders/\\1D8DgOdjT35Vf5Tqej3K5ZWFqz3LhgeQt?usp=sharing}.

\section{Conclusion}
\label{sec:conclusion}

CAM-based methods have seen large success in still images for WSOL task. Due to several limitations in current work in WSVOL task, we propose to leverage CAMs for this task. However, since CAMs are not designed to benefit from temporal information in videos, we propose a new module, CAM-TMP, that allows CAMs to do so. It aims to collect available ROI from a sequence of CAMs, which are used to generate pseudo-labels for training. Combined with local and global constraints, we are able to train our model for WSVOL task. Evaluated on two public benchmarks for unconstrained videos, we demonstrated that simple CAM-methods can yield competitive results. Our method yielded new state-of-the-art localization performance. Our ablations show that localization improvement in our method can be done by leveraging short time dependency. Demonstrative videos suggest that our proposal can be easily adapted for subsequent tasks such as video object tracking and detection.

\section*{Acknowledgment}
This research was supported in part by the Canadian Institutes of Health Research, the Natural Sciences and Engineering Research Council of Canada, and the Digital Research Alliance of Canada (alliancecan.ca).


\appendices

%
%
%
%

\section{Visual results}
\label{sec:visual-results}

\begin{figure}
     \centering
     \begin{subfigure}[b]{0.45\textwidth}
         \centering
         \includegraphics[width=\textwidth]{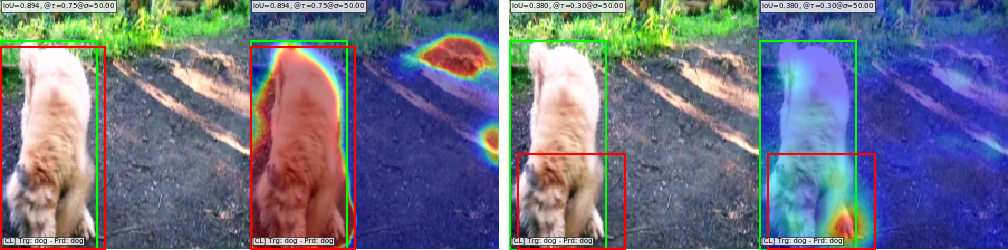}
     \end{subfigure}
     \begin{subfigure}[b]{0.45\textwidth}
         \centering
         \includegraphics[width=\textwidth]{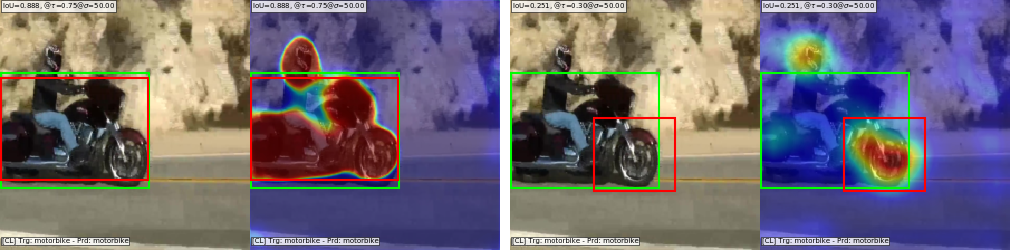}
     \end{subfigure}
     \begin{subfigure}[b]{0.45\textwidth}
         \centering
         \includegraphics[width=\textwidth]{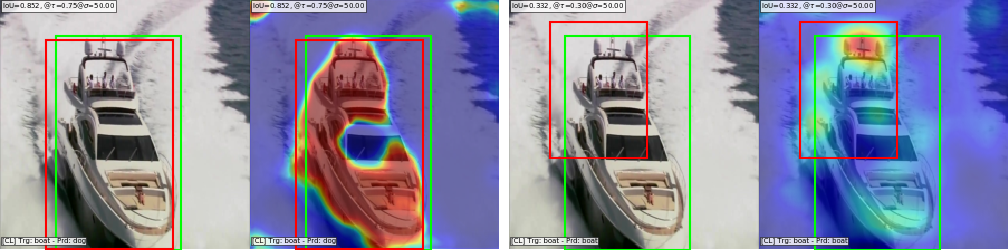}
     \end{subfigure}
     \begin{subfigure}[b]{0.45\textwidth}
         \centering
         \includegraphics[width=\textwidth]{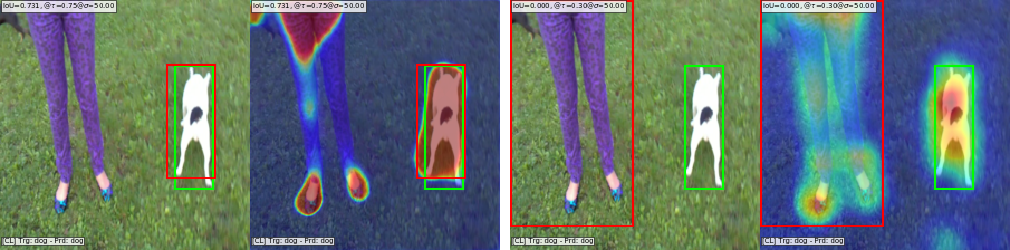}
     \end{subfigure}
     \begin{subfigure}[b]{0.45\textwidth}
         \centering
         \includegraphics[width=\textwidth]{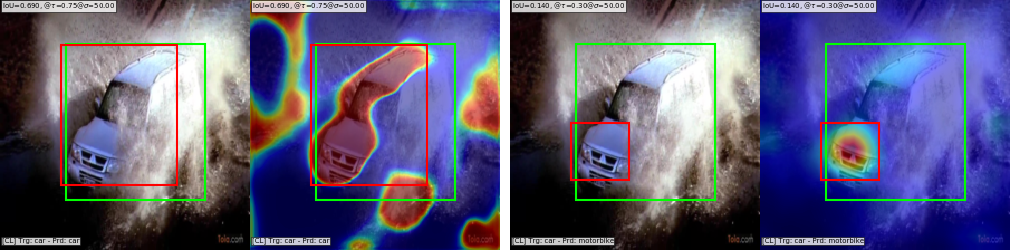}
     \end{subfigure}
     \begin{subfigure}[b]{0.45\textwidth}
         \centering
         \includegraphics[width=\textwidth]{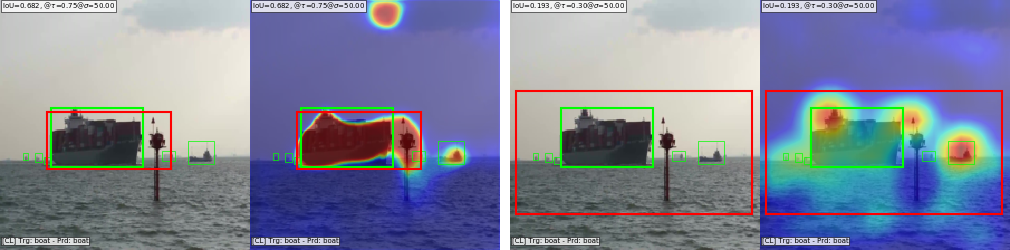}
     \end{subfigure}
     \begin{subfigure}[b]{0.45\textwidth}
         \centering
         \includegraphics[width=\textwidth]{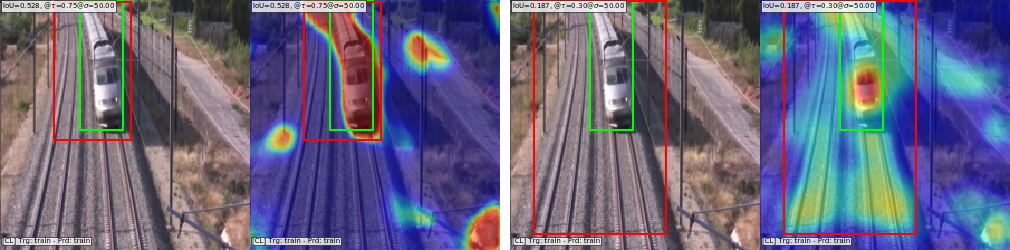}
     \end{subfigure}
     \begin{subfigure}[b]{0.45\textwidth}
         \centering
         \includegraphics[width=\textwidth]{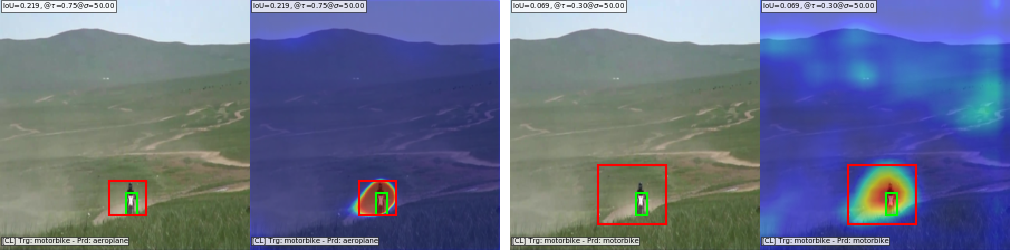}
     \end{subfigure}
     \begin{subfigure}[b]{0.45\textwidth}
         \centering
         \includegraphics[width=\textwidth]{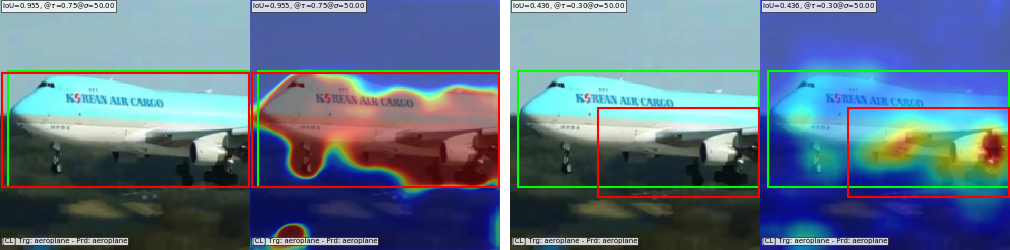}
     \end{subfigure}
        \caption{Prediction examples of test sets frames. \emph{Left}: TCAM (ours). \emph{Right}: baseline CAM method, LayerCAM~\citep{JiangZHCW21layercam}. \emph{Bounding box}: ground truth (green), prediction (red). Second column is predicted CAM over image.}
        \label{fig:pred-sm-1}
\end{figure}

\begin{figure}
     \centering
     \begin{subfigure}[b]{0.45\textwidth}
         \centering
         \includegraphics[width=\textwidth]{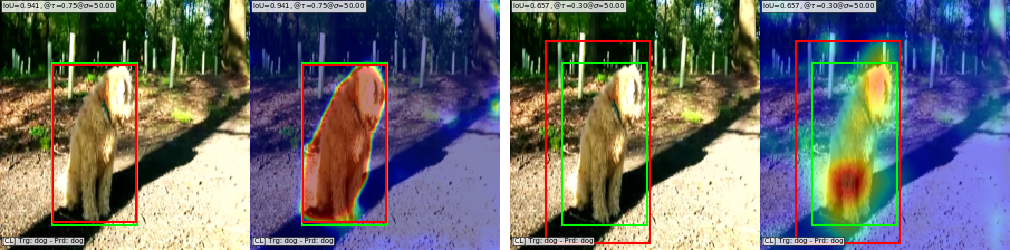}
     \end{subfigure}
     \\
     \begin{subfigure}[b]{0.45\textwidth}
         \centering
         \includegraphics[width=\textwidth]{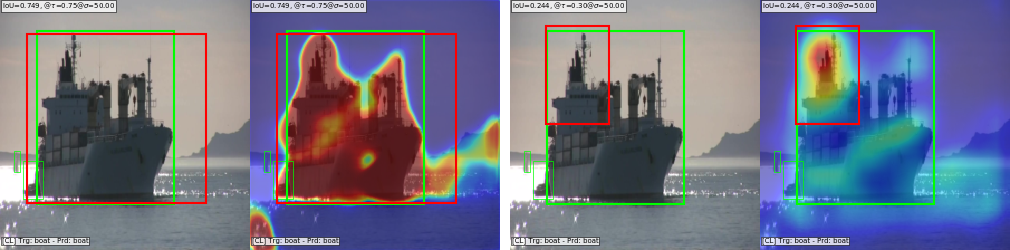}
     \end{subfigure}
     \\
     \begin{subfigure}[b]{0.45\textwidth}
         \centering
         \includegraphics[width=\textwidth]{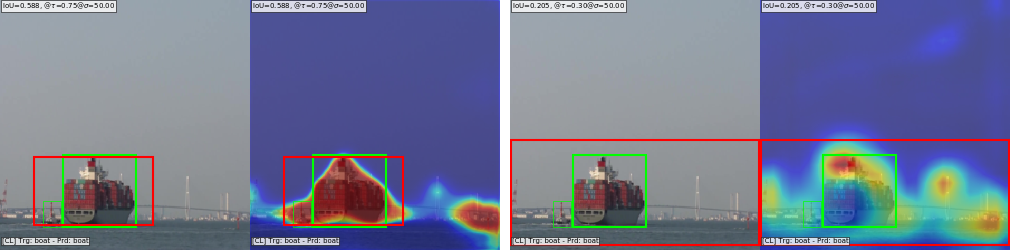}
     \end{subfigure}
     \\
     \begin{subfigure}[b]{0.45\textwidth}
         \centering
         \includegraphics[width=\textwidth]{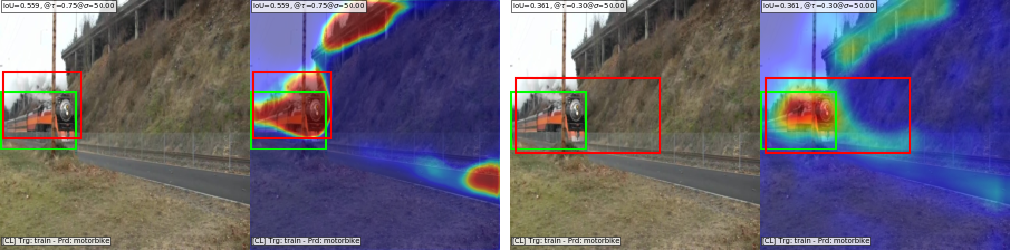}
     \end{subfigure}
     \\
       \begin{subfigure}[b]{0.45\textwidth}
         \centering
         \includegraphics[width=\textwidth]{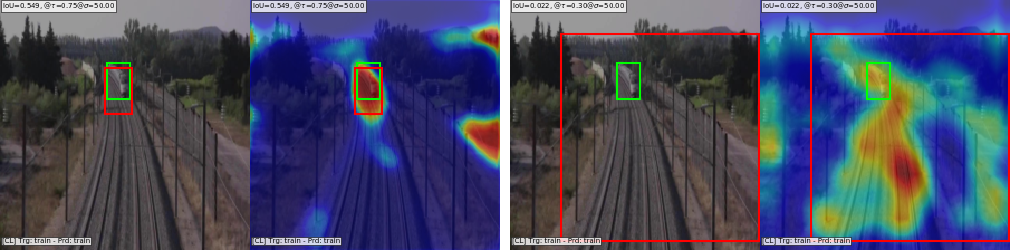}
     \end{subfigure}
     \\ %
     \begin{subfigure}[b]{0.45\textwidth}
         \centering
         \includegraphics[width=\textwidth]{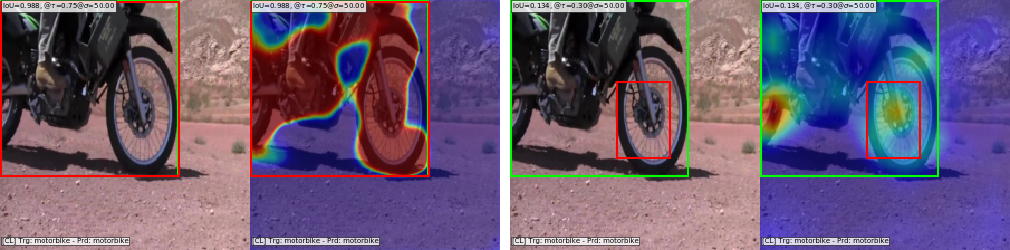}
     \end{subfigure}
     \\
     \begin{subfigure}[b]{0.45\textwidth}
         \centering
         \includegraphics[width=\textwidth]{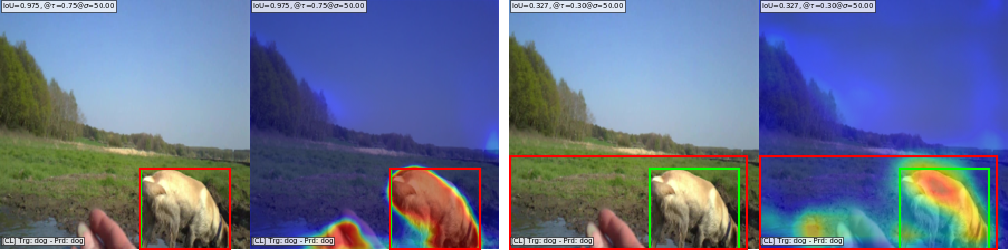}
     \end{subfigure}
     \begin{subfigure}[b]{0.45\textwidth}
         \centering
         \includegraphics[width=\textwidth]{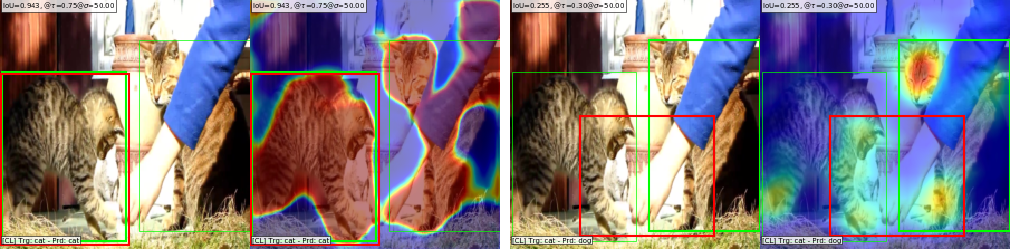}
     \end{subfigure}
     \begin{subfigure}[b]{0.45\textwidth}
         \centering
         \includegraphics[width=\textwidth]{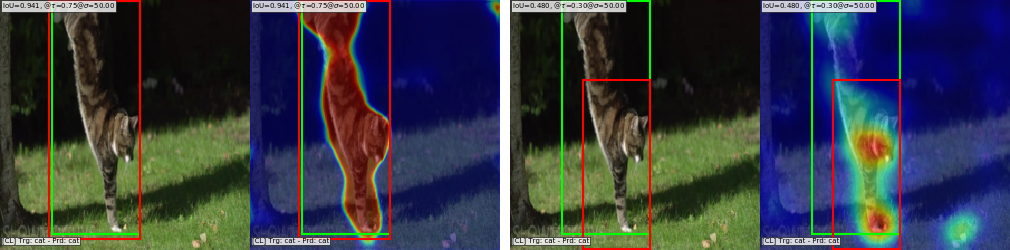}
     \end{subfigure}
        \caption{Prediction examples of test sets frames. \emph{Left}: TCAM (ours). \emph{Right}: baseline CAM method, LayerCAM~\citep{JiangZHCW21layercam}. \emph{Bounding box}: ground truth (green), prediction (red). Second column is predicted CAM over image.}
        \label{fig:pred-sm-2}
\end{figure}

Fig.\ref{fig:pred-sm-1}, \ref{fig:pred-sm-2} present more prediction cases over labeled ground truth frames.

\FloatBarrier

\bibliographystyle{apalike}
\bibliography{biblio}

\end{document}